\documentclass[10pt,twocolumn,letterpaper]{article}
\pdfoutput=1

\usepackage[pagenumbers]{cvpr}      

\usepackage{graphicx}
\usepackage{amsmath}
\usepackage{amssymb}
\usepackage{booktabs}
\usepackage{soul}
\usepackage{caption}
\usepackage{subcaption}
\usepackage{enumitem}
\usepackage[pagebackref,breaklinks,colorlinks]{hyperref}

\usepackage[capitalize]{cleveref}
\crefname{section}{Sec.}{Secs.}
\Crefname{section}{Section}{Sections}
\Crefname{table}{Table}{Tables}
\crefname{table}{Tab.}{Tabs.}
\Crefname{algocf}{Algorithm}{Algorithms}
\crefname{algocf}{Alg.}{Algs.}

\def\cvprPaperID{}
\def\confName{}
\def\confYear{}

\usepackage{siunitx}
\usepackage[ruled,linesnumbered]{algorithm2e}
\SetAlgoCaptionSeparator{.}
\SetAlCapSty{small}
\DontPrintSemicolon
\usepackage{collcell}
\usepackage{xfrac}
\usepackage{inconsolata}
\usepackage{mathtools}

\renewcommand{\vec}[1]{\boldsymbol{#1}}
\DeclareMathOperator{\diag}{diag}
\DeclareMathOperator*{\argmin}{argmin}

\newcommand{\cW}{\mathcal{W}}
\newcommand{\cB}{\mathcal{B}}
\newcommand{\cR}{\mathcal{R}}
\newcommand{\cL}{\mathcal{L}}
\newcommand{\cC}{\mathcal{C}}
\newcommand{\bW}{\mathbf{W}}
\newcommand{\bU}{\mathbf{U}}
\newcommand{\bS}{\mathbf{S}}
\newcommand{\bs}{\mathbf{s}}
\newcommand{\bV}{\mathbf{V}}
\newcommand{\bR}{\mathbf{R}}
\newcommand{\bL}{\mathbf{L}}
\newcommand{\bbl}{\mathbf{b}}

\SetKwComment{Comment}{$\triangleright$\ }{}
\SetCommentSty{emph}
\SetKwInput{KwHyperparams}{Hyperparameters}
\SetKwFunction{FTrain}{Train-NeRF}
\SetKwFunction{FTrainR}{STrain-NeRF}
\SetKwFunction{FCombine}{Combine}
\SetKwFunction{FParameterise}{Reparameterise}
\SetKwFunction{FReparameterise}{Refactorise}
\SetKwFunction{FRecover}{Recover}
\SetKw{KwFor}{for}
\newcommand{\MarkTransfer}{\textcolor{blue}{\ensuremath{\ast}}}
\newcommand{\MarkLowerBetter}{\textcolor{black}{\ensuremath{\blacktriangledown}}}
\newcommand{\MarkHigherBetter}{\textcolor{black}{\ensuremath{\blacktriangle}}}
\newcommand{\datasetname}[1]{{\ttfamily #1}}

\usepackage{flushend}

\begin{document}
\title{Federated Neural Radiance Fields}

\author{Lachlan Holden, Feras Dayoub, David Harvey, Tat-Jun Chin\\
Australian Institute for Machine Learning\\
The University of Adelaide, SA 5005, Australia\\
{\tt\small \{lachlan.holden,feras.dayoub,david.harvey,tat-jun.chin\}@adelaide.edu.au}
}
\maketitle


\begin{abstract}
The ability of neural radiance fields or NeRFs to conduct accurate 3D modelling has motivated application of the technique to scene representation. Previous approaches have mainly followed a centralised learning paradigm, which assumes that all training images are available on one compute node for training. In this paper, we consider training NeRFs in a federated manner, whereby multiple compute nodes, each having acquired a distinct set of observations of the overall scene, learn a common NeRF in parallel. This supports the scenario of cooperatively modelling a scene using multiple agents. Our contribution is the first federated learning algorithm for NeRF, which splits the training effort across multiple compute nodes and obviates the need to pool the images at a central node. A technique based on low-rank decomposition of NeRF layers is introduced to reduce bandwidth consumption to transmit the model parameters for aggregation. Transferring compressed models instead of the raw data also contributes to the privacy of the data collecting agents.
\end{abstract}

\section{Introduction}

Neural radiance fields (NeRF)~\cite{mildenhall.etal.2020_nerf} are a recently developed neural network model that are proving to be a powerful tool for 3D modelling. Trained on a finite set of image observations of a scene with ground truth camera poses, the output of a NeRF can be used to synthesise photo-realistic images from previously unseen viewpoints of the scene. While originally applied to object-centric scenes, there is increasing work on using NeRFs to model outdoor environments\cite{turki.etal.2022_meganerf,tancik.etal.2022_blocknerf,xiangli.etal.2022_bungeenerf}. This could contribute to capabilities such as augmented reality and autonomous navigation.

\begin{figure}[t]\centering
     \begin{subfigure}[b]{0.99\columnwidth}
         \centering
         \includegraphics{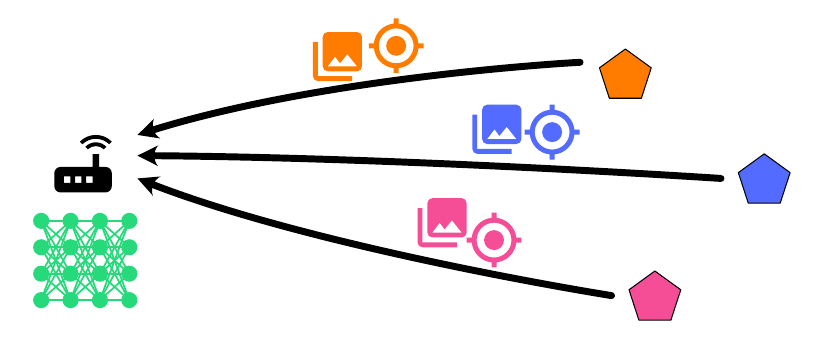}
         \caption{Centralised learning: all client data is transferred to a server before training of the NeRF can take place.}
         \label{fig:centralised}
     \end{subfigure}
     \begin{subfigure}[b]{0.99\columnwidth}
         \centering
         \includegraphics{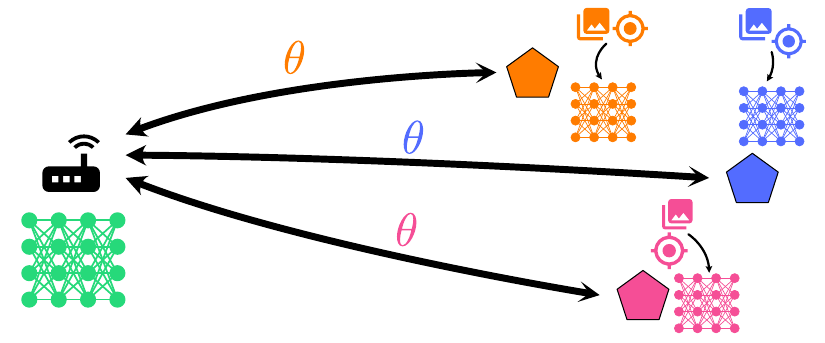}
         \caption{Federated learning: our proposed FedNeRF splits the training effort across the clients, each using only its own data on board. Only the network weights are transferred between client and server for aggregation.}
         \label{fig:distributed_with_fed}
     \end{subfigure}     
    \caption{Centralised versus federated learning of NeRF.}
    \label{fig:centralised_vs_distributed}
\end{figure}

Existing NeRF methods mostly take a \emph{centralised learning} paradigm, which requires all training images to be available on a single compute node before the training of NeRF can proceed. On the other hand, many researchers have advocated using multiple agents (\eg, tourists, cars, UAVs) to cooperatively collect data from outdoor environments~\cite{mcenroe.etal.2022_survey,doan.etal.2019_scalable,martin-brualla.etal.2021_nerf}. Employing this data collection approach to environment modelling using NeRF entails transferring all collected images to a central node (see Fig.~\ref{fig:centralised}), which can incur high data communication costs. Moreover, the privacy of the data collectors is not guaranteed due to the sharing of the data to a central pool.

As edge computing devices increase in processing capability~\cite{murshed.etal.2021_machine,mcenroe.etal.2022_survey}, it has become practical to perform neural network optimisation on the edge. If the effort to train an overall model can be distributed and parallelised across multiple edge devices, particularly if each device needs to take into account only the data that is on board, the potential to speed up training convergence of the overall model is significant. In addition, the need to pool the data is alleviated.

Towards collaborative sensing and learning of NeRF for scene modelling, the main question we ask in this paper is: \emph{Can a NeRF be learned in a parallel and distributed manner across multiple compute nodes, without pooling the data?}

\vspace{-1em}
\paragraph{Contributions.}

We answer the question in the affirmative by developing a novel federated learning algorithm for NeRF, called \textbf{FedNeRF}. Our experiments validate that the structure of traditional MLP-based NeRF is amenable to optimisation across multiple compute nodes using standard federated averaging techniques, where each node tunes the model using an independent set of image observations of the overall scene. To reconcile the resultant models at the different nodes, only the optimised weights need to be communicated to a central node for aggregation into an overall NeRF; see Fig.~\ref{fig:distributed_with_fed}.

To reduce bandwidth consumption during the update cycles of FedNeRF, we introduce a simple model reparametrisation technique based on low-rank approximation to reduce the number of effective parameters in NeRF. Results show that this compresses \emph{each} weight update by as much as 65\% (resulting in total bandwidth reduction of 97.5\%) without greatly affecting the accuracy of the learned NeRF, indicating a high level of redundancy in the model.

\vspace{-1em}
\paragraph{Benefits.}

FedNeRF is able to leverage multiple compute nodes to speed up training convergence of NeRF. In the context of multi-agent sensing and learning, FedNeRF incurs smaller data communication costs due to the low-rank compression applied on the NeRF weights. Transferring the compressed models and avoiding data sharing and pooling also help maintain the privacy of the data collection agents.

\section{Related work}

NeRF is a relatively new technique~\cite{mildenhall.etal.2020_nerf} that is receiving a lot of attention in the vision community. Here, we survey recent advances that are relevant to the proposed FedNeRF.

\subsection{NeRF}

A NeRF is a neural network that implicitly represents a 3D scene in that a NeRF can accurately render arbitrary views of the scene. A NeRF model is a much more compact (\eg, 5 MB~\cite{mildenhall.etal.2020_nerf}) than storing all the training images or other conventional 3D model representations. More technical details of NeRF will be presented in Sec.~\ref{sec:prelim}.

To improve the robustness of NeRF against imaging conditions (\eg, lighting, exposure) and transient occlusions (\eg, pedestrians, cars), NeRF in the Wild (NeRF-W)~\cite{martin-brualla.etal.2021_nerf} includes a learned appearance embedding for each image for the colour-generating part of the network. Transient occlusions are handled by adding a new transient head to the model with a second learned embedding that generates an uncertainty field alongside a transient colour and density.

Standard NeRF works well for object-centric and indoor scenes but is less capable of modelling outdoor scenes with far-away backgrounds. NeRF++~\cite{zhang.etal.2020_nerf} addresses this by using two NeRFs; one modelling the foreground objects in a unit sphere, and the other using an inverted sphere reparameterisation to represent everything outside the sphere. 

DeRF \cite{rebain.etal.2021_derf}, and KiloNeRF \cite{reiser.etal.2021_kilonerf} also decompose a NeRF into multiple sub-NeRFs, but to improve the rendering time. DeRF uses a modest number of sub-networks split using learned Voronoi decomposition, while KiloNeRF uses a large number of small, efficient NeRFs in a spatial grid. 

\subsection{Large-scale NeRF}\label{sec:largescale_survey}

Scaling up NeRF to represent large scenes (\eg, streetscapes) while retaining fine details is an active research topic with potential applications in localisation and navigation~\cite{moreau.etal.2021_lens,yen-chen.etal.2021_inerf}. It has been found that simply increasing the network size has diminishing returns on the quality of synthesised images for more complex scenes \cite{rebain.etal.2021_derf}. More successful approaches conduct spatial or geometric decomposition to model a large scene using a collective of NeRFs.

To model a city block, Block-NeRF trains a set of NeRFs distributed in and between city block intersections~\cite{tancik.etal.2022_blocknerf}. Each NeRF is trained on the set of training images with camera pose within a certain radius of some training area origin. At inference time, the final image is created by combining the rendered images of each NeRF whose training area includes the query pose. Instead of partitioning the training data by full image based on the pose of the camera as in Block-NeRF, in Mega-NeRF~\cite{turki.etal.2022_meganerf} each \emph{pixel} is included in the train sets for each NeRF whose area its ray intersects. At inference time, the rendered image is created by combining the outputs of each NeRF whose area is intersected by the ray of each pixel. Instead of separating NeRFs spatially, BungeeNeRF creates progressively larger models with segmented training data of multiple scales to be able to render high-quality outputs in larger scenes at a range of distances and resolutions~\cite{xiangli.etal.2022_bungeenerf}. It is based on Mip-NeRF \cite{barron.etal.2021_mipnerf}, which uses conical frustums rather than rays to represent pixels to better render fine details at high resolutions.

\paragraph{Contrasting FedNeRF with large-scale NeRF.}

We stress that FedNeRF solves a task that is orthogonal to that addressed by large-scale NeRF techniques~\cite{tancik.etal.2022_blocknerf,turki.etal.2022_meganerf,xiangli.etal.2022_bungeenerf}. While large-scale NeRFs decompose the \emph{scene} into separate cells where each is modelled by an individual NeRF, FedNeRF distributes the \emph{data} (or \emph{data collection effort}) and \emph{model optimisation} over multiple clients.

Combining FedNeRF with large-scale NeRF techniques is possible, which we leave as future work.

\subsection{Federated learning}

Federated learning is a mechanism by which multiple agents cooperating can train a single neural network in a shared way \cite{konecny.etal.2015_federated,mcmahan.etal.2017_communicationefficient,mcenroe.etal.2022_survey}. There is a single central server node and a number of client nodes. Training is performed distributed over the clients on data collected by each client, and this data does \emph{not} need to be sent to the server -- instead, only network weights are transferred back and forth.

Due to the need to transfer network weights, reducing bandwidth usage in federated learning is important. Techniques include distributed optimisation~\cite{mills.etal.2020_communicationefficient}, client dropout \cite{wen.etal.2022_federated,bouacida.etal.2021_adaptive}, update compression via techniques like sparsification and quantisation \cite{mills.etal.2020_communicationefficient,aji.heafield.2017_sparse,lin.etal.2022_deep,sattler.etal.2019_sparse,sattler.etal.2020_robust,xu.etal.2022_ternary}, and intermediate update aggregation \cite{chen.etal.2020_communicationefficient}. Another technique, and the one adopted by our work, involves reparameterising the model to yield fewer effective weights \cite{chen.etal.2022_update,konecny.etal.2017_federated}.

While federated learning is a mature topic, we are the first to adapt it to NeRF and achieve concrete results in collaborative multi-agent sensing and modelling.

\section{Preliminaries}\label{sec:prelim}

Here, we provide a brief introduction to NeRF; the reader is referred to~\cite{mildenhall.etal.2020_nerf} for more details. In its original form, a NeRF is a fully-connected neural network with intermittent skip connections that map a 5D input, consisting of a 3D point in the scene $\vec x = (x, y, z)$ and the pitch and yaw of a viewing direction $\vec d = (\beta, \phi)$, into a density $\sigma$ and RGB colour $\vec c = (r, g, b)$. A synthetic image of the scene can be rendered through traditional volume rendering techniques by sampling colour and opacity from the network at numerous points along rays through the scene.

A NeRF first predicts $\sigma$ solely from $\vec x$, and then this output is concatenated with $\vec d$ and fed to the second part of the network to predict $\vec c$. Additionally, two networks are actually trained simultaneously – a coarse and fine network pair. The output of the coarse is used during rendering to inform the samples taken from the fine network. To enable the networks to represent high-frequency features in the scenes, the inputs are first transformed by a positional encoding.

The parameters of a NeRF can be defined as $\theta = \{\mathcal{W}, \mathcal{B} \}$, where $\mathcal{W} = \{ \bW_1, \bW_2, \dots \}$ are matrices representing the weights of the fully connected (FC) layers, and $\mathcal{B} = \{ \vec b_1, \vec b_2, \dots \}$ are the corresponding bias vectors. Note that for a single NeRF, these sets include the parameters for both the coarse and fine networks together.

A NeRF is trained on a dataset $D = \{(I_\ell, p_\ell)\}^{N}_{\ell=1}$ consisting of images $I_\ell$ of a scene and corresponding poses $p_\ell$. In each iteration of the training, a batch of pixels are randomly sampled from a selected image $I_\ell$. For each pixel ray, the network is queried as in inference to render a pixel colour, and the mean squared error (MSE) between this colour and the ground-truth pixel colour is employed as the training loss to update the network parameters $\theta$.

\section{Application scenario}\label{sec:scenario}

\begin{figure*}[ht]\centering
    \begin{subfigure}{0.33\textwidth}\centering
        \includegraphics{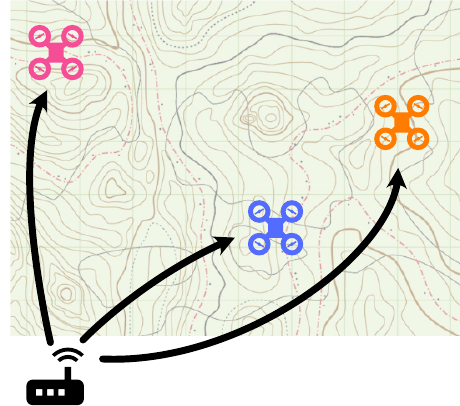}
        \caption{Receive initial weights from the server.}
    \end{subfigure}
    \begin{subfigure}{0.33\textwidth}\centering
        \includegraphics{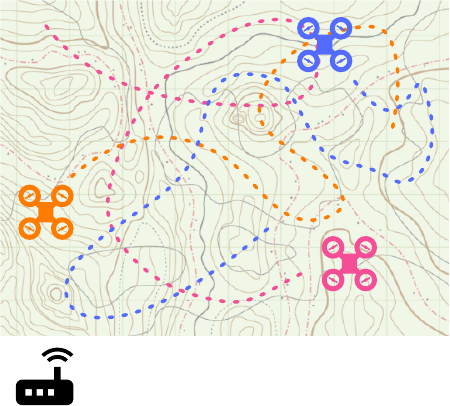}
        \caption{Take new photos and update individual NeRFs.}
    \end{subfigure}
    \begin{subfigure}{0.33\textwidth}\centering
        \includegraphics{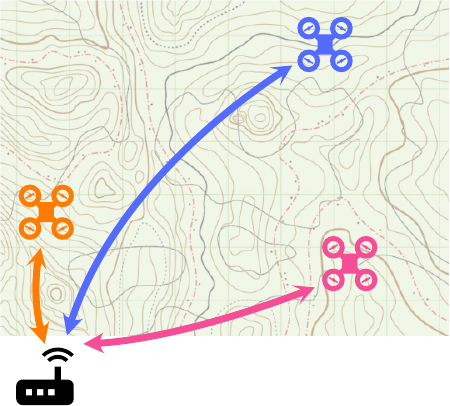}
        \caption{Aggregate weights with server.}
    \end{subfigure}
    \caption{Main steps of FedNeRF, framed in an example application scenario of mapping an outdoor area with a fleet of quadcopters.}
    \label{fig:federated-learning}
\end{figure*}

Consider $K$ agents (a.k.a.~clients) that work collaboratively to map a scene. The clients are equipped with sensors and edge compute devices, and can communicate with a central server. The clients collect images of the scene with ground truth poses derived from onboard positioning systems; this requirement is realistic for platforms such as vehicles and UAVs (\cf~\cite{mcenroe.etal.2022_survey,doan.etal.2019_scalable,turki.etal.2022_meganerf}). Let $D_{k}$ be the data collected by the $k$-th client. The overall aim is to learn a NeRF $\theta$ using all the data collected $\{ D_{k} \}^K_{k=1}$. \cref{fig:federated-learning} illustrates.

For brevity, the scenario description considers only one data collection round. However, the methods in the rest of the paper can be executed sequentially to handle multiple rounds, \eg, to conduct lifelong mapping~\cite{doan.etal.2019_scalable,warburg.etal.2020_mapillary}. 

\vspace{-1em}
\paragraph{Baseline method.}
Given initial NeRF weights $\theta^{(0)}$ (\eg, the estimate on previously collected data or on an arbitrarily chosen $D_k$), the baseline method (\cref{alg:naive}) involves each client sending the recorded data back to the server. The initial $\theta^{(0)}$ is then refined at the server using all collected data.

The bandwidth consumption of the baseline method is
\begin{equation}\label{eq:B-naive}
    B_\text{baseline} = \sum_k \; \lvert D_k \rvert \text{,}
\end{equation}
where $\lvert D_k \rvert$ is the size of data $D_k$ as the number of bytes. In practice, $B_\text{baseline}$ is significant to due transferring raw images (even with JPEG compression). Moreover, the baseline method does not fully exploit the compute power of the clients, and can lead to slower convergence than distributed optimisation, as we will show in \cref{sec:results}.

\begin{algorithm}
\caption{Baseline method for the application scenario (\cref{sec:scenario}). Data transfers are marked by \MarkTransfer.}\label{alg:naive}
\KwHyperparams{
\begin{itemize}[itemsep=0pt,parsep=0pt,topsep=2pt]
    \item Number of training iterations $T$
\end{itemize}
}
$\theta^{(0)}$ are initial NeRF weights \;
\For(\Comment*[f]{parallel loop}){client $k \gets 1,\ldots,K$}{
    $D_k$ is data collected by client $k$ \;
    $D_k$ is transmitted to the server \MarkTransfer \;
}
$\theta^{(1)} \gets$ \FTrain{$\theta^{(0)}$, $\{D_k\}_{k=1,\dots,K}, T$} \Comment{standard NeRF training; see \cref{sec:prelim}}
\KwRet $\theta^{(1)}$
\end{algorithm}

\section{FedNeRF}\label{sec:our-approach}

The proposed FedNeRF is summarised in \cref{alg:fednerf} and illustrated in \cref{fig:distributed_with_fed,fig:federated-learning}. The main distinguishing features of FedNeRF from the baseline method are:
\begin{itemize}[leftmargin=1em,itemsep=0pt,parsep=0pt,topsep=2pt]
    \item The sharing of NeRF weights to each client which are then refined onboard using collected data.
    \item The transfer of the individual refined weights back to the server to be aggregated into an overall model.
    \item The compression of updates through a low-rank model reparameterisation technique to reduce bandwidth consumption due to the transfer of the weights.
\end{itemize}
It is vital to note that the \textbf{for} loop in Steps~\ref{step:forstart} to \ref{step:forend} is \emph{executed in parallel} on the clients. Also, \cref{alg:fednerf} can be executed sequentially for lifelong mapping. The rest of this section will provide more details of the major steps in \cref{alg:fednerf}.

\begin{algorithm}
\caption{Proposed FedNeRF for the application scenario (Sec.~\ref{sec:scenario}). Data transfers are marked by \MarkTransfer. }\label{alg:fednerf}
\KwHyperparams{
\begin{itemize}[itemsep=0pt,parsep=0pt,topsep=2pt]
    \item Variance $\alpha$
    \item Number of merge rounds $M$
    \item Number of training iterations per merge $\Upsilon$
\end{itemize}}
$\theta^{(0)} = \{\mathcal{W}^{(0)},\mathcal{B}^{(0)}\}$ are initial NeRF weights \;
$\cL^{(0)}, \cR \gets$ \FParameterise{$\cW^{(0)},\alpha$} \Comment{these are the learnable and frozen parameters; see \cref{sec:update-compression}} \label{step:reparam} 
$\cC^{(0)} \gets \{ \cL^{(0)}, \cB^{(0)} \} $ \;
$\cR$ is distributed to all clients $k = 1,\dots,K$ \MarkTransfer \;
\For(\Comment*[f]{parallel loop}){client $k \gets 1,\dots,K$}{
    $D_k$ is data collected by client $k$
}
\For{merge round $m = 1, \ldots, M$}{
\For(\Comment*[f]{parallel loop}){client $k = 1,\dots,K$}{ \label{step:forstart}    
        $\cC^{(m-1)}$ is received from the server \MarkTransfer \;
        $\cC^{(m)}_k \gets$ \FTrainR{$\cC^{(m-1)}$, $\cR$, $D_k$, $\Upsilon$} \Comment{$\cC^{(m)}_k$ contains $\cL_k^{(m)},\cB^{(m)}_k$; see \cref{sec:update-compression}} \label{step:strain}
        $\cC^{(m)}_k$ is transmitted to the server \MarkTransfer \;
    } \label{step:forend}
    \For{$k = 1,\dots,K$}{    \label{step:recoverstart}
        $\cW^{(m)}_k \gets$ \FRecover{$\cL_k^{(m)}, \cR$} \Comment{see \cref{sec:update-compression}}
        $\theta^{(m)}_k \gets \{ \cW^{(m)}_k, \cB^{(m)}_k \} $ \;
    } \label{step:recoverend}
    $\theta^{(m)} \gets$ \FCombine{$\{\theta_k^{(m)}\}_{k=1,\dots,K}$} \label{step:combine} \Comment{federated averaging; see \cref{sec:fedavg}}
    $\cL^{(m)} \gets$ \FReparameterise{$\cW^{(m)}, \cR$}  \label{step:refactor} \Comment{extract learnable parameters; see \cref{sec:update-compression}}
    $\cC^{(m)} \gets \{ \cL^{(m)}, \cB^{(m)} \} $ \;
}
\KwRet $\theta^{(M)}$
\end{algorithm}

\subsection{Federated averaging for NeRF}\label{sec:fedavg}

Inspired by federated learning/averaging algorithms~\cite{mcmahan.etal.2017_communicationefficient,zhu.etal.2021_federated}, FedNeRF conducts $M$ merging rounds to combine the models that are individually optimised by the $K$ clients into an overall model. In each merging round, given $K$ individual NeRF models $\theta_k = \{ \cW_k, \cB_k \}$, $k = 1,\dots,K$, where
\begin{align}
    \cW_k = \{ \bW_{k,1}, \bW_{k,2}, \dots \}, \;\;\; \cB_k = \{ \bbl_{k,1}, \bbl_{k,2}, \dots \},
\end{align}
we aim to compute a merged model $\tilde{\theta} = \{ \tilde{\cW}, \tilde{\cB} \}$, where
\begin{align}
    \tilde{\cW} = \{ \tilde{\bW}_{1}, \tilde{\bW}_{2}, \dots \}, \;\;\; \tilde{\cB} = \{ \tilde{\bbl}_{1}, \tilde{\bbl}_{2}, \dots \}.
\end{align}
The combination step (Step~\ref{step:combine}) in \cref{alg:fednerf} obtains the merged weights by computing for each NeRF layer $z$
\begin{equation}\label{eq:fed-avg}
    \tilde{\bW}_{z} = \frac{\sum_{k = 1}^K  \lvert D_k \rvert \bW_{k,z} }{\sum_{k = 1}^K \lvert D_k \rvert},
\end{equation}
where $|D_k|$ is the size of the data collected by the $k$-th client. A similar weighted averaging is conducted for the bias parameters $\tilde{\cB}$. The merged model $\tilde{\theta}$ is then redistributed to each client for further refinement and merging.

\subsection{Update compression for NeRF}\label{sec:update-compression}

In place of transferring the training images from the clients to the server, \cref{alg:fednerf} conducts multiple exchanges of NeRF weights. Thus, it is vital to compress the amount of data that is transmitted in FedNeRF to be competitive in terms of bandwidth consumption. To this end, we adopt the update compression method of~\cite{chen.etal.2022_update} to NeRF.

\subsubsection{Reparameterisation}

The first step (Step~\ref{step:reparam} in \cref{alg:fednerf}) is to linearly factorise the FC layers in the initial model. Via singular value decomposition (SVD), the $z$-th FC layer $\bW_z \in \mathbb{R}^{u \times v}$ becomes
\begin{equation}
    \bW_z = \bU_z \bS_z \bV_z^\top \text{,}
\end{equation}
where $\bU_z \in \mathbb R^{u\times u}$ contains the left singular vectors, $\bV_z \in \mathbb R^{v \times v}$ contains the right singular vectors, and $\bS_z \in \mathbb R^{u \times v}$ is a diagonal matrix that contains the singular values, \ie,
\begin{align}
    \bS_z = \diag(\bs_z)
\end{align}
with vector $\bs_z$ of length $\min(u,v)$. Then, $\bW_z$ is truncated to rank $r$, where $r \le \min(u,v)$, by computing
\begin{equation}
    \bW^r_z = \underbrace{\bU_{z,:,1:r} \diag(\bs_{z,1:r})}_{\bL_z} \underbrace{\bV_{z,:,1:r}^\top}_{\bR_z} \text{,}
\end{equation}
where $\bU_{z,:,1:r}$ are the first-$r$ columns of $\bU_z$ (similarly for $\bV_{z,:,1:r}$), and $\bs_{z,1:r}$ are the first-$r$ elements of $\bs_z$. Following~\cite{chen.etal.2022_update}, $\bL_z \in \mathbb{R}^{u \times r}$ becomes the \emph{learnable} weights, while $\bR_z \in \mathbb{R}^{r \times v}$ are \emph{frozen}. Performing the rank-$r$ truncation on all FC layers $z = 1,2,\dots$, we have
\begin{align}
    \cL = \{ \bL_1, \bL_2, \dots \}, \;\;\; \cR = \{ \bR_1, \bR_2, \dots \}.
\end{align}
The frozen parameters $\cR$ are shared with all clients prior to merging. Only $\cL$ (along with corresponding biases $\cB$) will be iteratively refined and exchanged between the server and clients during merging.

The value of $r$ is guided by the amount of variance $\alpha$ to be retained in $\bW^r_z$. Specifically, we pick the smallest $r$ s.t.
\begin{align}
    \frac{\sum_{i=1}^r s_{z,i} }{ \sum_{i=1}^{\min(u,v)} s_{z,i} } \ge \alpha, \label{eq:alpha}
\end{align}
where $s_{z,i}$ is the $i$-th singular value of $\bW_z$. This implies that different FC layers in the NeRF will be truncated to different ranks. \Cref{sec:experiments} will discuss the selection of $\alpha$.

\subsubsection{Sparse training}

Reparameterisation is conducted only once on the server side, then the learnable parameters are progressively updated by the clients through the rest of FedNeRF. In Step~\ref{step:strain} of \cref{alg:fednerf}, client $k$ refines the NeRF using data $D_k$ by training for $\Upsilon$ rounds on images $I \in D_k$ in a similar way as in \cref{sec:prelim}, except the training loss is backpropagated through the learnable parameters $\cC_k$ to update them only, leaving the frozen parameters $\cR$ unchanged.

\subsubsection{Model recovery and re-factorisation}

After the clients individually update the learnable parameters, the individual NeRFs are recovered
\begin{align}
    \cC_k = \{ \cL_{k}, \cB_{k} \} \mapsto \theta_{k} = \{ \cW_{k}, \cB_{k} \}
\end{align}
in Steps~\ref{step:recoverstart} to~\ref{step:recoverend} by recomputing each FC layer as
\begin{align}
    \bW_{k,z} = \bL_{k,z} \bR_z.
\end{align}
After federated averaging in Step~\ref{step:combine}, the combined NeRF is refactorised to extract updated $\cL = \{\tilde\bL_1, \tilde\bL_2,\ldots\}$ in Step~\ref{step:refactor} by solving for each FC layer
\begin{align}
    \tilde{\bL}_z = \argmin_{\Gamma \in \mathbb{R}^{u \times r}} \| \tilde{\bW}_z - \Gamma \bR_z \|_F,
\end{align}
which can be achieved via a linear solver. Alternatively, by recognising that the process~\eqref{eq:fed-avg} is linear, the FC layers of the merged NeRF are already factorisable by $\cR$ by default. Thus, Steps~\ref{step:combine} and \ref{step:refactor} for the weights can be simultaneously achieved by averaging the updated learnable layers
\begin{align}
    \tilde{\bL}_z = \frac{\sum_{k = 1}^K  \lvert D_k \rvert \bL_{k,z} }{\sum_{k = 1}^K \lvert D_k \rvert}.
\end{align}

\subsection{Bandwidth usage}

Considering the lines marked with \MarkTransfer{} in \cref{alg:fednerf}, the total amount of data transferred in FedNeRF is 
\begin{align}\label{eq:B-federated}
        B_\text{FedNeRF} = K(|\cR| + 2M |\cC|),
\end{align}
where $\lvert \cR \rvert$ and $\lvert \cC \rvert$ are respectively the size in bytes of the frozen and learnable parameters of the NeRF model. The sizes can be reduced using standard file zipping methods.

The compression ratio CR achieved by FedNeRF over the baseline method is thus
\begin{equation}
    \text{CR} = B_\text{baseline} / B_\text{FedNeRF} \text{,}
\end{equation}
where higher CR means more economical bandwidth utilisation by FedNeRF. Of course, CR is inversely related to variance $\alpha$, which in turn is directly related to the representation power of the NeRF. \cref{sec:experiments} will examine the trade-off.

\section{Experiments}\label{sec:experiments}
We conduct a set of experiments to demonstrate that NeRFs can be learned in a federated manner. This section describes our experimental setup and details the datasets, evaluation metrics and implementation.

\subsection{Experimental setup}\label{sec:exp-setup}

\textbf{Datasets.} First, our technique is applied to a range of the real (\datasetname{horns}, \datasetname{fern}) and synthetic (\datasetname{lego}, \datasetname{drums}, \datasetname{materials}, \datasetname{ship}) single-object scenes used in the original NeRF paper \cite{mildenhall.etal.2020_nerf}. This is done to validate that the structure of NeRFs is amenable to this federated training with update compression. To demonstrate the data compression capabilities of FedNeRF, we use a set of \num{1000} training images generated at random poses from the \datasetname{lego} scene, dubbed the \datasetname{lego\_xl} dataset. Note that the choice of 1000 images is relatively arbitrary -- the idea is to use this as a proof-of-concept larger dataset from readily-available data to demonstrate that compression is possible. As such, the compression ratio results would vary significantly with different image counts. Finally, to investigate the performance of FedNeRF in an outdoor scene, we use the \datasetname{building} dataset from Mega-NeRF~\cite{turki.etal.2022_meganerf}.

\textbf{Evaluation metrics.} Following \cite{mildenhall.etal.2020_nerf}, we use the following metrics: image loss, learned perceptual image patch similarity (LPIPS) \cite{zhang.etal.2018_unreasonable}, structural similarity (SSIM) \cite{wang.etal.2004_image}, and peak signal-to-noise ratio (PSNR).

\textbf{Baselines.} For all of our experiments, an initial network $\theta^{(0)}$ is trained on a subset of the data. Given $\theta^{(0)}$, FedNeRF (\cref{alg:fednerf}) and the baseline (\cref{alg:naive}) were both executed and compared. This is especially an appropriate experimental setting for multi-round/lifelong learning campaigns, which is a key potential application of FedNeRF.

\textbf{Implementation.} For the synthetic scenes, all initial $\theta^{(0)}$ were trained for \num{20000} iterations on \sfrac{1}{5} of the original data, chosen randomly. The baseline method was then trained for $T = \num{20000}$ more iterations on the remaining \sfrac{4}{5}. FedNeRF was trained with update compression similarly from $\theta^{(0)}$ for $M = 20$ and $\Upsilon = \num{1000}$ with $K=4$ clients, each receiving \sfrac{1}{4} of the remaining \sfrac{4}{5}. The images for the clients were split such that the views were interspersed for these proof-of-concept experiments -- further investigation into non-IID partitioning is left as future work. The same setting was followed for the real scenes, except with $T=\num{25000}$ for initial and baseline, and $M=25$ and $\Upsilon = \num{1000}$ for FedNeRF. In each experiment, $\alpha = 90\%$.

The settings above mean that the number of iterations \emph{per client} for FedNeRF was the same as the number of iterations for the baseline \ie, that $M \Upsilon = T$. While this meant that the cumulative iterations across all clients $KM\Upsilon$ of FedNeRF was higher than the baseline's $T$, the setting was justified since the clients ran in parallel. Of course each client operated on only a quarter of the data available to the baseline. In any case, we have tested the setting where $KM\Upsilon = T$, which yielded only slightly lower accuracy for FedNeRF.

For the \datasetname{lego\_xl} scene, three experiments as above were run by varying $\alpha$ and $M$ -- one maximising network quality with minimal compression ($\alpha=90\%,M=20$), one maximising bandwidth compression ($\alpha=75\%,M=4$), and one balanced experiment in between ($\alpha=80\%,M=10$).

Similar experiments to the \datasetname{lego\_xl} dataset were undertaken for the \datasetname{building} dataset. A single cell was trained on a section of the scene using Mega-NeRF's spatial partitioning regime. The network structure was similar to that of standard NeRF, with the addition of per-image learned appearance embedding that were client specific. The training set for the single cell included all pixels from the whole scene which intersect the cell's area. The initial model $\theta^{(0)}$ was trained for \num{20000} iterations on 320 images, with baseline being trained for a further $T = \num{80000}$ iterations. For FedNeRF, $M$ was varied (160, 80, 20) with $\Upsilon = T/M$, and $K=2$ with each client having 800 images.

\begin{figure}[tp] \centering
    \includegraphics{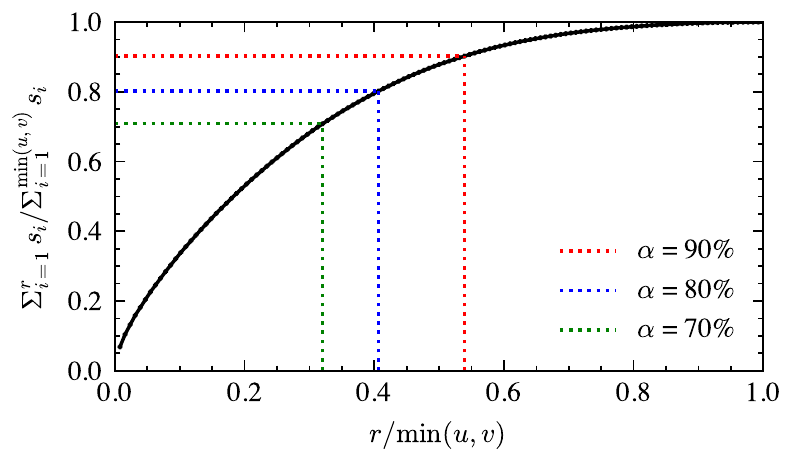}
    \caption{Representative plot of normalised cumulative sum of singular values $s$ against $r$ for layers in a NeRF. Most of the fully-connected layers exhibit this elbow-shaped characteristic, while a small number are closer to a linear relation. $\alpha$ is as in \cref{eq:alpha}.}
    \label{fig:cumsum-s-r}
\end{figure}

\textbf{Hyperparameter settings.} The hyperparameters that affect the structure of the NeRF networks (number of layers, dimensions of layers, etc.) were selected to be similar to those used in the original papers \cite{mildenhall.etal.2020_nerf,turki.etal.2022_meganerf}.

The graph in \cref{fig:cumsum-s-r} informed the selection of $\alpha$. Setting $\alpha$ to 70\%--90\% could produce large savings in parameter counts for most layers. Initial experiments indicated that values of $\alpha < 70\%$ start to cause training to fail.

The same optimisation configuration is used for all experiments as in the original papers. The optimisers' states are reset after training of the initial network for both baseline and FedNeRF. The optimiser is not reset for each training round between merges for the federated experiments.

Our focus was not to beat the synthesis quality of NeRF and Mega-NeRF as reported in~\cite{mildenhall.etal.2020_nerf,turki.etal.2022_meganerf} (and as we are not altering the network structure, we do not believe this to be possible), but instead show that in some cases FedNeRF was able to converge faster than the baseline for a given number of training iterations. As such, the training iteration counts here were lower than what is used traditionally, but sufficient to produce good-quality images and allow for comparison of the techniques.

\subsection{Results and discussion}\label{sec:results}

\begin{table*}[tp]
    \caption{Validation results for federated learning of real-world single object scenes. Lower/higher better is indicated by \MarkLowerBetter/\MarkHigherBetter{} respectively.
    }
    \label{tab:nerf}
    \centering
    \begingroup\small
\begin{tabular}[t]{@{}>{\collectcell{\datasetname}}l<{\endcollectcell}cccccccccccc@{}}
    \toprule
     & \multicolumn{3}{c}{Loss \MarkLowerBetter} & \multicolumn{3}{c}{LPIPS \MarkLowerBetter} & \multicolumn{3}{c}{SSIM \MarkHigherBetter} & \multicolumn{3}{c}{PSNR \MarkHigherBetter} \\
    \cmidrule(lr){2-4} \cmidrule(lr){5-7} \cmidrule(lr){8-10} \cmidrule(l){11-13}
    \multicolumn{1}{l}{Scene} & Init. & Base. & Fed. & Init. & Base. & Fed. & Init. & Base. & Fed. & Init. & Base. & Fed. \\
    \midrule
    fern & 0.0096 & 0.0055 & 0.0055 & 0.4951 & 0.4756 & 0.4841 & 0.5832 & 0.6406 & 0.6426 & 20.18 & 22.58 & 22.64 \\
    horns & 0.0048 & 0.0046 & 0.0039 & 0.5092 & 0.4862 & 0.4836 & 0.6520 & 0.6653 & 0.6836 & 23.20 & 23.41 & 25.36 \\
    \bottomrule
\end{tabular}
\endgroup
\end{table*}
\begin{figure*}[tp]\centering
    \includegraphics{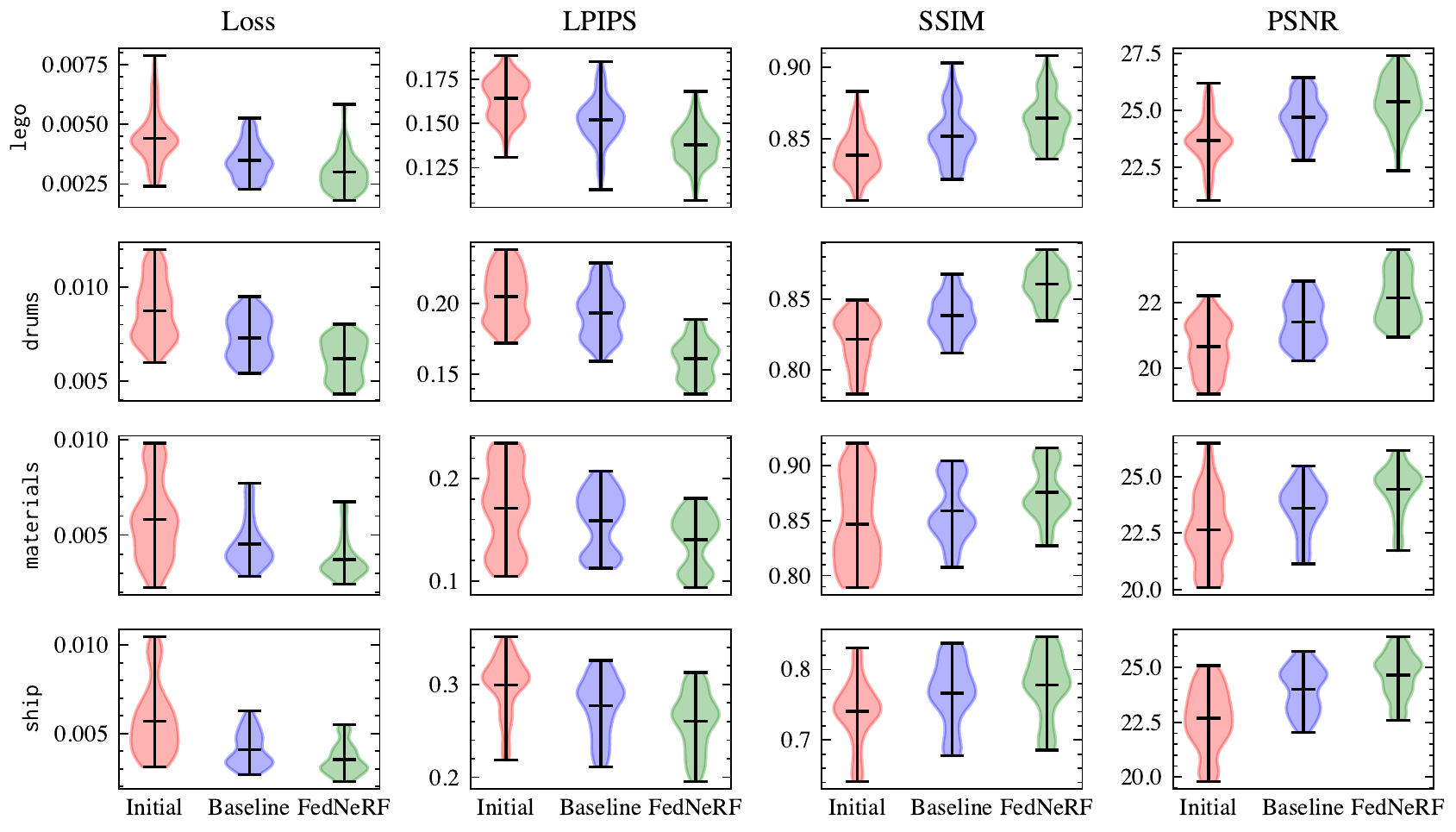}    
    \caption{Validation results for the federated learning of synthetic single-object NeRF scenes. For loss and LPIPS lower is better, and for SSIM and PSNR higher is better. Our method is shown in green, compared to the baseline approach shown in blue.}
    \label{fig:object-scenes}
\end{figure*}

\begin{figure*}[!thp]\centering
    \includegraphics{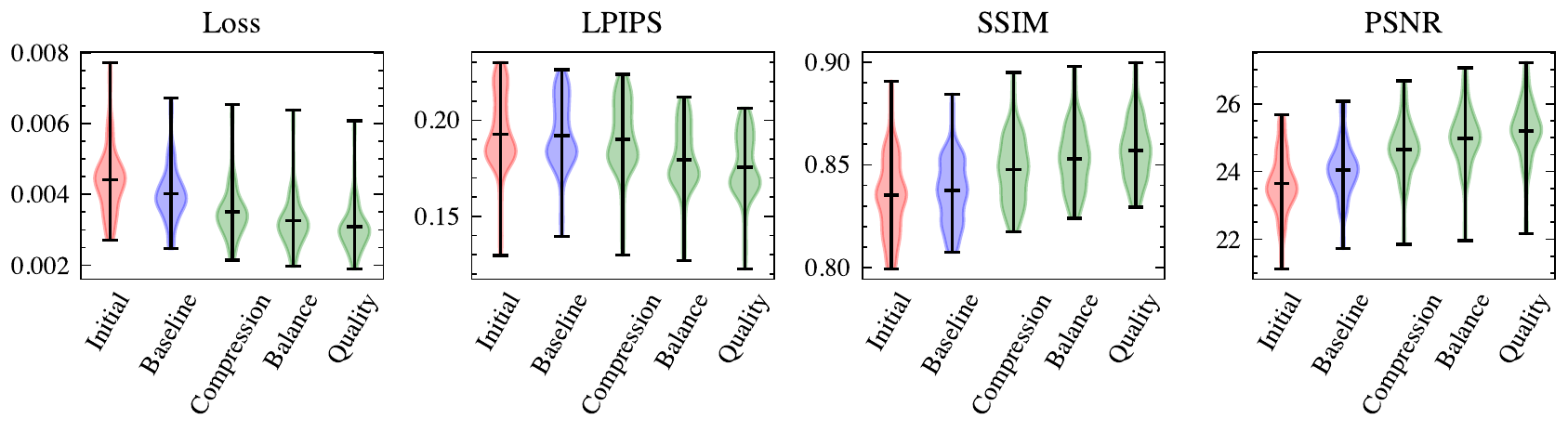}
    \caption{Validation results for the federated learning of higher image count synthetic NeRF scene \datasetname{lego\_xl}. For loss and LPIPS lower is better, and for SSIM and PSNR higher is better. Our method is shown in green, compared to the baseline approach shown in blue.}
    \label{fig:lego-xl}
\end{figure*}

\begin{table*}[!tpb]
    \caption{Mean validation results for federated learning of a single cell of the \datasetname{building} Mega-NeRF scene. Lower/higher better is indicated by \MarkLowerBetter/\MarkHigherBetter{} respectively. The dataset already uses JPEG-encoded images, so the compression ratios presented include this factor. $\alpha$ is \% variance retained as in \eqref{eq:alpha}. $M$ is number of merges. CR is compression ratio.}
    \label{tab:mega-nerf}
    \centering
    \begingroup\small
\begin{tabular}{@{}lrrrccccccccc@{}}
    \toprule
    & & & & \multicolumn{3}{c}{LPIPS \MarkLowerBetter} & \multicolumn{3}{c}{SSIM \MarkHigherBetter} & \multicolumn{3}{c}{PSNR \MarkHigherBetter} \\
     \cmidrule(lr){5-7} \cmidrule(lr){8-10} \cmidrule(l){11-13}
    Experiment & $\alpha$ & $M$ & CR & Init. & Base. & Fed. & Init. & Base. & Fed. & Init. & Base. & Fed. \\
    \midrule
    Quality & 90\% & 160 & 1.00 & 0.7274 & 0.6532 & 0.7057 & 0.3370 & 0.3871 & 0.3608 & 17.14 & 18.58 & 17.89 \\
    Balance & 80\% & 80 & 2.39 & & & 0.7398 & & & 0.3564 & & & 17.71 \\
    Compression & 75\% & 20 & 10.33 & & & 0.7389 & & & 0.3558 & & & 17.72\\
    \bottomrule
\end{tabular}
\endgroup
\end{table*}

\begin{table}[htbp]
    \caption{Compression ratio results for federated learning of higher image count synthetic NeRF scene \datasetname{lego\_xl}. JPEG compression with 90\% quality is used on the original PNG images, resulting in images c.\ \sfrac14 the size. $\alpha$ is \% variance retained as in \eqref{eq:alpha}. $M$ is number of merges.}
    \label{tab:nerf-xl-comp}
    \centering
    \begingroup\small
\begin{tabular}{@{}lrrrr@{}}
    \toprule
    & & & \multicolumn{2}{c@{}}{Compression Ratio} \\
    \cmidrule(l){4-5}
    Experiment & $\alpha$ & $M$ & PNG & JPEG \\
    \midrule
    Quality & 90\% & 20 & 5.96 & 1.49 \\
    Balance & 80\% & 10 & 15.29 & 3.82 \\
    Compression & 75\% & 4 & 40.11 & 10.03 \\
    \bottomrule
\end{tabular}
\endgroup
\end{table}

\begin{figure}[!htb]
    \centering
    \includegraphics{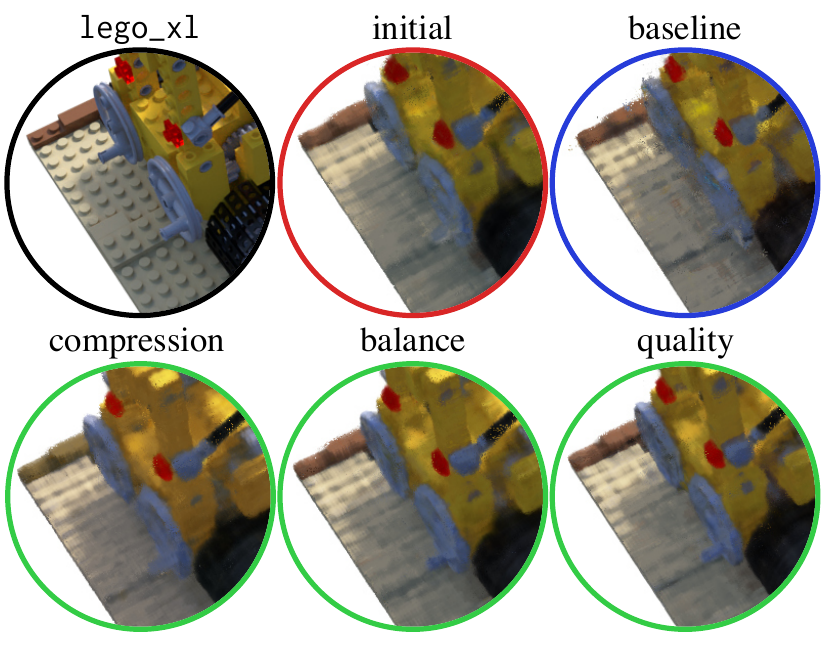}
    \caption{Close-up qualitative results for the single object \datasetname{lego\_xl}, demonstrating the improved results of FedNeRF.}
    \label{fig:qualitative-close-up}
\end{figure}

The results in \cref{tab:nerf} for the real-world object scenes with one validation image each and \cref{fig:object-scenes} for the synthetic scenes with 200 validation images each show that the internal structure of these NeRFs are capable of being learned in a federated way, and are amenable to update compression. In fact, for every metric for every scene other than LPIPS for the \datasetname{fern} scene, our FedNeRF method both improved on the initial network and outperformed the baseline network, which demonstrates that our method results in faster convergence than centralised training for these circumstances.

\begin{figure}[!htb]
    \centering
    \includegraphics{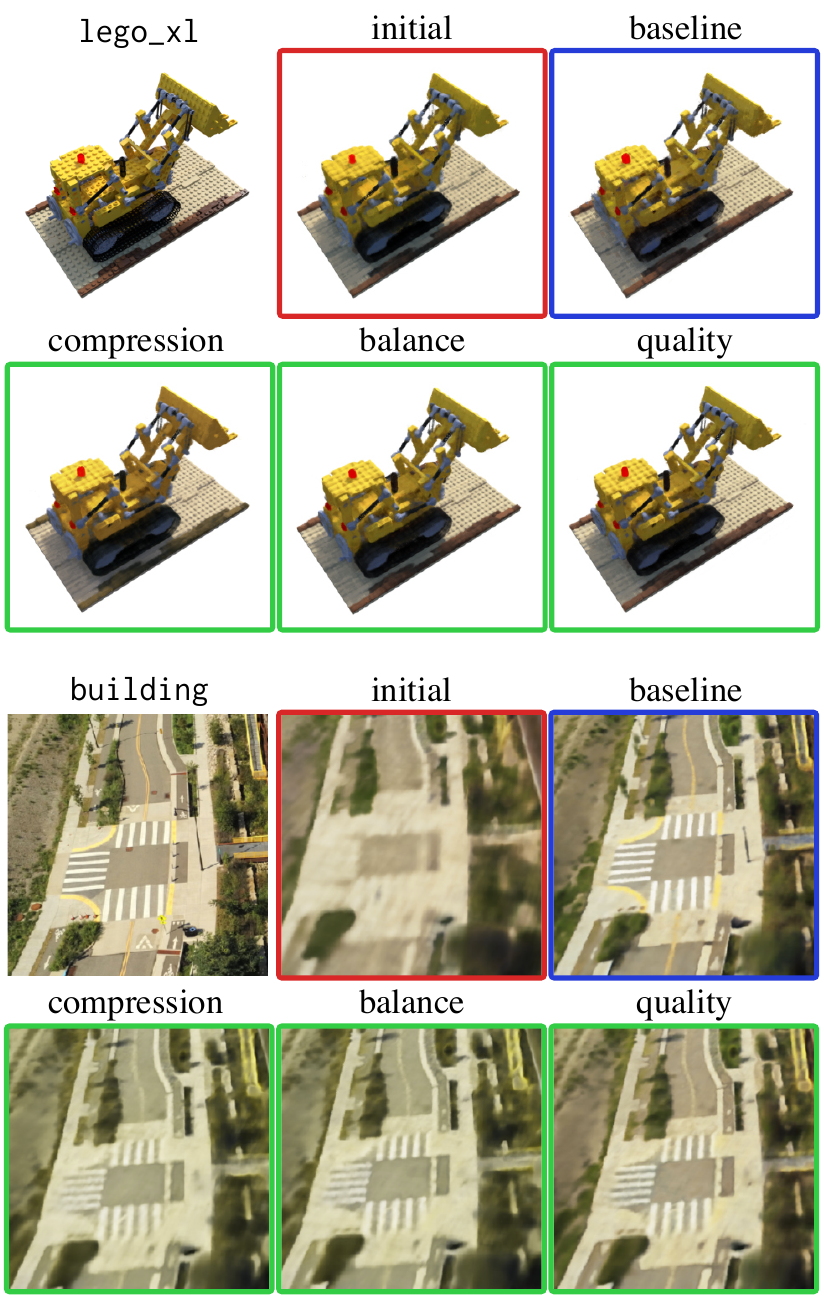}
    \caption{Qualitative results for the single object \datasetname{lego\_xl} and outdoor \datasetname{building} scenes.}
    \label{fig:qualitative}
\end{figure}

The results presented in \cref{fig:lego-xl} and \cref{tab:nerf-xl-comp} on the \datasetname{lego\_xl} dataset show that FedNeRF both uses less bandwidth and achieves better quality faster than the baseline approach. The improved quality is clear in the qualitative results in \cref{fig:qualitative,fig:qualitative-close-up}.

In the outdoor scene case, our results in \cref{tab:mega-nerf} that FedNeRF consistently shows improvements over the initial network, but in contrast to the previous results, it does not converge faster than the baseline network. Examining the qualitative results shown in \cref{fig:qualitative} highlight the significance of the improvements over the initial network, though, where the outputs of FedNeRF are much more suitable to a range of tasks like mapping and exploration. Additionally, FedNeRF still exhibits significant bandwidth savings, and potential privacy improvements, and makes better use of the clients' compute via parallelisation.

\section{Conclusions and future work}

In this paper, we present FedNeRF, a novel federated learning algorithm to train a NeRF in a parallel and distributed manner across multiple compute nodes without pooling the data. Our results show that the structure of NeRF is amenable to optimisation across multiple compute nodes, where each node tunes the model using an independent set of image observations of the overall scene. The update compression method in FedNeRF was able to reduce the bandwidth consumption by more than 90\% in certain cases without affecting the accuracy of the resulting NeRF.

These experiments demonstrate that FedNeRF works as a proof-of-concept, and valuable future work would involve testing more ``real-world'' experimental settings, including non-IID data partitioning, multiple data collection rounds, and more complex scenes with higher image counts.

\clearpage
{\small
\bibliographystyle{ieee_fullname}
\bibliography{zotero}
}

\end{document}